%% file: main.tex
\documentclass[11pt,a4paper]{article}

\usepackage[hyperref]{naaclhlt2018}
\usepackage{times}
\usepackage{latexsym}
\usepackage{url}
\usepackage{times}  
\usepackage{helvet}  
\usepackage{courier}  
\usepackage{url}  
\usepackage{graphicx}  
\usepackage{enumitem}
\usepackage{amsmath}
\usepackage{multirow}
\usepackage{siunitx} 
\usepackage{graphicx} 
\usepackage{caption}
\usepackage{subcaption}
\usepackage{tikz}
\usepackage{pgfplots}
\usepackage{color,soul}
\usepackage{amssymb}
\usepackage{tablefootnote}
\usepackage{bm}
\usepackage{booktabs}
\usepackage{xspace}
\usepackage{footnote}
\usepackage{empheq}
\usepackage{algorithm}
\usepackage[noend]{algpseudocode}
\usepackage{adjustbox} 
\usepackage{array}
\usepackage{etoolbox} 
\newcommand{\ubold}{\fontseries{b}\selectfont} 
\robustify\ubold

\aclfinalcopy

\newcommand{\poorya}[1]{{\color{black} #1}}
\newcommand{\reza}[1]{{\color{black} #1}}
\newcommand*\unk{\textsc{$<$Unk$>$}\xspace}
\newcommand*\sq{\textsc{Seq2Seq}\xspace}

\newcommand{\oRNN}{\operatorname{RNN}}
\newcommand{\softmax}{\operatorname{softmax}}
\newcommand{\dlstm}{\operatorname{disLSTMs}}
\newcommand{\srep}{\operatorname{shrRep}}

\newcommand{\va}{\pmb{a}}
\newcommand{\vE}{\pmb{E}}
\newcommand{\vtheta}{\pmb{\theta}}

\definecolor{blau_2b}{RGB}{0,131,204}

\input{notation.tex}

\usepackage{tabularx}
\usepackage{threeparttable}

\title{Neural Machine Translation for Bilingually Scarce Scenarios: \\ A Deep Multi-task Learning Approach}

\author{Poorya Zaremoodi \hspace{1.5cm}  Gholamreza Haffari\\
  Faculty of Information Technology,
   Monash University, Australia \\ 
  {\tt first.last@monash.edu} \\}

\date{}
\begin{document}

\maketitle
\begin{abstract}

Neural machine translation requires large amounts of parallel training text
to learn a reasonable-quality translation model. 
This is particularly inconvenient for language pairs for which enough parallel text is not available. 
In this paper, we use monolingual linguistic resources in the source side to 
address this challenging problem based on a multi-task learning approach. 
More specifically, we {scaffold}  the machine translation task on auxiliary tasks including 
semantic parsing, syntactic parsing, and named-entity recognition. 
This effectively injects  semantic and/or syntactic knowledge into the translation model, 
which would otherwise require a large amount of training bitext.
We empirically evaluate and show the effectiveness of our multi-task learning approach on three translation tasks: 
English-to-French, English-to-Farsi, and English-to-Vietnamese. 
\end{abstract}

\section{Introduction}
\label{sec::in}
\input{sec10-intro}

\section{Neural  \sq Transduction}
\label{sec::am}
\input{sec20-nmt}

\section{\sq Multi-Task Learning }
\label{sec::multitask}
\input{sec30-multitask}
\section{Adversarial Training}
\input{sec40-adv-train.tex}

\section{Experiments}
\label{sec::experiments}
\input{sec60-experiments}

\subsection{Analysis}
\input{sec70-analysis}

\section{Related Work}
\input{sec80-related}

\section{Conclusions and Future Work}
\input{sec90-conc.tex}

\section*{Acknowledgments}
The research reported here was initiated at the 2017 Frederick Jelinek Memorial Summer Workshop on Speech and Language Technologies, hosted at Carnegie Mellon University and sponsored by Johns Hopkins University with unrestricted gifts from Amazon, Apple, Facebook, Google, and Microsoft.
We are very grateful to the workshop members for the insightful discussions and  data pre-processing. 
This work was supported by computational resources from the Multi-modal Australian ScienceS Imaging and Visualisation Environment (MASSIVE) at Monash University. 
This work was supported by the Australian Research Council via DP160102686.
The first author was partly supported by CSIRO's Data61.

\bibliography{refs}
\bibliographystyle{acl_natbib}

\end{document}

%% file: notation.tex
\DeclareMathOperator*{\argmax}{arg\,max}

\newcommand{\vv}{\boldsymbol{v}}

\newcommand{\vh}{\boldsymbol{h}}
\newcommand{\vx}{\boldsymbol{x}}
\newcommand{\vy}{\boldsymbol{y}}
\newcommand{\vr}{\boldsymbol{r}}

\newcommand{\vs}{\boldsymbol{s}}

\newcommand{\vW}{\boldsymbol{W}}
\newcommand{\vb}{\boldsymbol{b}}
\newcommand{\vc}{\boldsymbol{c}}
\newcommand{\valpha}{\boldsymbol{\alpha}}

\newcommand{\vTheta}{\boldsymbol{\Theta}}

%% file: sec10-intro.tex
Neural Machine Translation (NMT) with attentional encoder-decoder architectures \cite{luong-pham-manning:2015:EMNLP, bahdanau2014neural} has revolutionised machine translation, and  achieved state-of-the-art for several language pairs. 
However, NMT is notorious for its need \reza{for} large amount\reza{s} of bilingual data \cite{DBLP:journals/corr/KoehnK17} to achieve reasonable translation quality. Leveraging existing monolingual  resources is a potential approach for compensating this requirement in bilingually scarce scenarios. 
Ideally,  semantic and syntactic knowledge learned from existing linguistic resources provides NMT with proper inductive biases, leading to increased generalisation and better translation quality.

Multi-task learning (MTL) is an effective approach to inject knowledge \reza{into a}   task, which is learned from other related tasks. 
Various recent works have attempted to improve NMT with an MTL approach  \cite{peng2017deep,liu2017adversarial,zhang2016exploiting}; however, they either do not make use of curated linguistic resources \cite{domhan2017using,zhang2016exploiting}, or their MTL architectures are restrictive yielding mediocre improvements \cite{niehues2017exploiting}. The current research leaves open  how to best leverage curated linguistic resources in a suitable MTL framework to improve  NMT.

In this paper, we make use of curated monolingual linguistic resources in the source side to improve NMT in bilingually scarce scenarios. 
More specifically, we {scaffold}  the machine translation task on auxiliary tasks including 
semantic parsing, syntactic parsing, and named-entity recognition. 
This is achieved by casting the auxiliary tasks as \reza{sequence-to-sequence} (\sq) transduction tasks,
and tie the parameters of their encoders and/or decoders with those of the main translation task.  
Our MTL architectures makes use of deep stacked encoders and decoders, where the parameters of the top layers are shared across the tasks. 
We further make use of adversarial training  to prevent contamination of common knowledge with \reza{task-specific } information. 

We present  empirical results \reza{on} translating from English into French, Vietnamese, and Farsi; three target languages with varying degree of  divergence compared to English. 
Our extensive empirical results demonstrate the effectiveness of our MTL approach in substantially improving the translation quality for these three translation tasks in bilingually scarce scenarios.

%% file: sec20-nmt.tex
Our MTL is based on the attentional encoder-decoder architecture for  \sq  transduction.  It contains an encoder to \emph{read} the input sentence, and an attentional decoder to \emph{generate} the output.

\paragraph{Encoder}
The encoder is a \reza{bi-directional} RNN whose hidden states represent tokens of the input sequence.  
These representations capture information not only of the corresponding token, but also other tokens in the sequence to leverage the context. 
The bi-directional RNN consists of two RNNs running in the left-to-right and right-to-left directions over the input sequence:
\begin{eqnarray}
\overrightarrow{\vh_{i}} &= \oRNN( \overrightarrow{\vh}_{i-1}, \vE_S [x_{i}]) \nonumber \\
\overleftarrow{\vh}_{i} &= \oRNN(\overleftarrow{\vh}_{i+1}, \vE_S [x_{i}]) \nonumber
\end{eqnarray}
where $\vE_S[x_i]$ is the embedding of the token $x_i$ from the embedding table $\vE_S$ of the input (source) space,  and $\overrightarrow{\vh}_{i}$ and $\overleftarrow{\vh}_{i}$ are the hidden states of the forward and backward RNNs which can be based on the LSTM (long-short term memory) \cite{Hochreiter:1997:LSM:1246443.1246450} or GRU (gated-recurrent unit) \cite{69e088c8129341ac89810907fe6b1bfe} units. Each source token is then represented by the concatenation of the corresponding bidirectional hidden states, $\vh_i=[\overrightarrow{\vh}_{i}; \overleftarrow{\vh}_{i}]$.

\paragraph{Decoder.}
The backbone of the decoder is a uni-directional RNN which generates the token of the output  one-by-one from left to right. The generation of each token $y_j$ is conditioned on all of the previously generated tokens $\vy_{<j}$ via the state of the RNN decoder $\vs_j$, and the input sequence via a \emph{dynamic} context vector $\vc_j$ (explained shortly):

\vspace{-5mm}
{\small
\begin{eqnarray}
\label{prob:seq2seq}
y_j &\sim& \softmax(\vW_{y}\cdot \vr_j+\vb_r)\\
\label{def:r}
\vr_j &=& \tanh(\vs_j + \vW_{rc}\cdot \vc_j +\vW_{rj}\cdot \vE_T[y_{j-1}])   \\
\vs_j &=& \tanh(\vW_{s}\cdot \vs_{j-1}+\vW_{sj}\cdot \vE_T[y_{j-1}]+\vW_{sc}\cdot \vc_j) \nonumber
\end{eqnarray}
}
where $\vE_T[y_j]$ is the embedding of the token $y_j$ from the embedding table $\vE_T$ of the output (target) space, and the $\vW$ matrices and $\vb_r$ vector are the parameters.

A crucial element of the decoder is the \emph{attention} mechanism which  dynamically attends to relevant parts of the input sequence necessary \reza{for} generating the next token \reza{in} the output sequence. Before generating the next token $t_j$, the decoder computes the attention vector $\valpha_j$ over the input token:
\begin{eqnarray*}
\valpha_j &=& \softmax(\va_j)\\
{a}_{ji}&=& \vv \cdot \tanh(\vW_{ae}\cdot \vh_i + \mathbf{W}_{at}\cdot \vs_{j-1})
\end{eqnarray*}
which intuitively is similar to the notion of \emph{alignment} in word/phrase-based statistical MT \cite{Brown:1993:MSM:972470.972474}. The attention vector is then used
to compute a fixed-length dynamic representation of the source sentence 
\begin{equation}
\label{def:c}
\vc_j =\sum_{i}\alpha_{ji}\vh_i.
\end{equation}
which is conditioned upon in the RNN decoder when computing the next state or generating the output word (as mentioned above). 

\paragraph{Training and Decoding.} The model parameters are trained end-to-end by maximising the (regularised) log-likelihood of the training data
\begin{equation*}
\arg\max_{\vtheta}\sum_{(\vx,\vy) \in \mathcal{D}} \sum_{j=1}^{|\vy|} \log P_{\vtheta}(y_j| \vy_{<j},\vx)
\end{equation*}
where the above conditional probability is  defined according to eqn (\ref{prob:seq2seq}). Usually drop-out is employed to prevent over-fitting on the training data. 
In the decoding time, the best output sequence for a given input sequence is produced by
\begin{equation*} 
 \arg\max_{\vy} P_{\vtheta}(\vy|\vx) =  \prod_j P_{\vtheta}(y_j|\vy_{<j}\vx).
 \end{equation*}
 Usually greedy decoding or beam search algorithms are employed to find an approximate solution, since solving the above optimisation problem exactly is computationally hard.

%% file: sec30-multitask.tex
We consider an extension of the basic \sq model where the encoder and decoder are equipped with deep stacked layers. Presumably, deeper layers capture more abstract information about a task, hence they can be used as a mechanism to share useful generalisable information among multiple tasks. 

\paragraph{Deep Stacked Encoder.} 
The deep encoder consists of multiple layers, where the hidden states \reza{in} layer $\ell-1$ are the inputs to the hidden states at the next layer $\ell$. That is,
\begin{eqnarray*}
\overrightarrow{\vh}_i^{\ell} = \overrightarrow{\oRNN}^{\ell}_{\vtheta_{\ell,enc}}(\overrightarrow{\vh}_{i-1}^{\ell},{\vh}_{i}^{\ell-1}) \\
\overleftarrow{\vh}_i^{\ell} = \overleftarrow{\oRNN}^{\ell}_{\vtheta_{\ell,enc}}(\overleftarrow{\vh}_{i-1}^{\ell},{\vh}_{i}^{\ell-1}) 
 \end{eqnarray*}
where ${\vh}_i^{\ell}=[\overrightarrow{\vh}_i^{\ell}  ;  \overleftarrow{\vh}_i^{\ell}]$ is the hidden state of the $\ell$'th layer RNN encoder for the $i$'th source sentence word. 
The inputs \reza{to} the first layer forward/backward RNNs  are the source word embeddings $\vE_S[x_i]$.
The representation of the source sentence is then the concatenation of the hidden states \reza{for} all layers
$
\vh_i = [\vh_i^1 ; \ldots ; \vh_i^{L}]
$
which is then used by the decoder.

\paragraph{Deep Stacked Decoder.}  
Similar to the multi-layer RNN encoder, the decoder RNN has multiple layers:
\begin{eqnarray*}
{\vs}_j^{\ell} = {\oRNN}^{\ell}_{\vtheta_{\ell,dec}}({\vs}_{j-1}^{\ell},{\vs}_{j}^{\ell-1}) 
 \end{eqnarray*}
where the inputs to the first layer RNNs are 
$$\vW_{sj}\cdot \vE_T[y_{j-1}]+\vW_{sc}\cdot \vc_j$$
 in which $\vc_j$ is the dynamic source context, as defined in eqn \ref{def:c}. 
 The state of the decoder  is then the concatenation of the hidden states \reza{for} all layers:
$
\vs_j = [\vs_j^1 ; \ldots ; \vs_j^{L}]
$
which is then used in eqn \ref{def:r} as part of the ``output generation module''. 

\paragraph{Shared Layer MTL.} 
We share the deep layer RNNs in the encoders and/or decoders across the tasks, as a mechanism to share abstract knowledge and increase model generalisation. 

Suppose  we have a total of $M+1$ tasks, consisting of the main task plus $M$ auxiliary tasks. 
Let $\vTheta_{enc}^m = \{\vtheta_{\ell,enc}^m\}_{\ell=1}^{L}$ and $\vTheta_{dec}^m =\{\vtheta_{\ell',dec}^m\}_{\ell'=1}^{L'}$ be the parameters of \reza{multi-layer} encoder and decoder for the task $m$. 
Let $\{\vTheta_{enc}^m,\vTheta_{dec}^m\}_{m=1}^M$ and $\{\vTheta_{enc}^0,\vTheta_{dec}^0\}$ be the RNN parameters \reza{for} the auxiliary tasks and the main task, respectively.
We share the parameters of the deep-level encoders and decoders of the auxiliary tasks with those of the main task.
That is,

\vspace{-5mm}
{\small
\begin{eqnarray*}
\forall m \in [1,..,M] \ \forall \ell \in [L_{enc}^m,.., L] &:&  \vtheta^m_{\ell,enc} = \vtheta^0_{\ell,enc}  \\
\forall m \in [1,..,M] \ \forall \ell' \in [{L'}_{dec}^m,.., L'] &:& \vtheta^m_{\ell',dec} = \vtheta^0_{\ell',dec} 
\end{eqnarray*}} 
where $L_{enc}^m$ and ${L'}_{dec}^m$ specify the deep-layer RNNs need to be shared parameters. 
Other parameters to share across the tasks include those of the attention module, the source/target embedding tables, and the output  generation module. 
As an extreme case, we can share \emph{all} the parameters of \sq architectures across the tasks. 

\paragraph{Training Objective.}  
Suppose we are given a collection of $M+1$ \sq transductions tasks,  each of which is associated with a training set $\mathcal{D}_m := \{(\vx_i,\vy_i)\}_{i=1}^{N_m}$. The parameters are learned by maximising  the MTL training objective: 
\begin{equation}
\label{mtl_obj}
\mathcal{L}_{mtl}(\vTheta_{mtl}) := \sum_{m=0}^M \frac{\gamma_m}{|\mathcal{D}_m|}\sum_{(\vx,\vy) \in \mathcal{D}_m}  \log P_{\vTheta_m}(\vy| \vx)
\end{equation}
where $\vTheta_{mtl}$ denotes all the parameters of the MTL architecture, $|\mathcal{D}_m|$ denotes the size of the training set for the task $m$, and $\gamma_m$ balances out its influence in the training objective. 

\paragraph{Training Schedule.} Variants of stochastic gradient descent (SGD)  can be used to optimise the objective in order to learn the parameters. 
Making the best use of tasks with different objective geometries is  challenging, e.g. due to the scale of their gradients.
One strategy for making an SGD update is to select the tasks from which the next data items should be chosen.
In our training schedule, we randomly select a training data item from the main task, and pair it with a data item selected from a randomly selected auxiliary task for making the next SGD update. 
This ensures the presence of training signal from the main task in all SGD updates, and avoids the training signal being washed out by the auxiliary tasks.

%% file: sec40-adv-train.tex
The learned shared knowledge can be contaminated by task-specific information.  We address this issue by adding \reza{an} adversarial objective.
The basic idea is to augment  the MTL training objective with additional terms, so that the identity of a task cannot be predicted from its data items by the representations \reza{resulted} from the shared encoder/decoder RNN layers. 

\paragraph{Task Discriminator.} The goal of the task discriminator is to predict the identity of a task for a data item based on the representations of the share layers.
More specifically, our task discriminator consists of two RNNs with LSTM units, each of which \reza{encodes} the sequence of hidden states in the shared layers of the encoder and the decoder.\footnote{When multiple layers are shared, we concatenate their hidden states at each time step, which is then input to the task discriminator's LSTMs.} The last hidden states of these two RNNs are then concatenated, giving rise to a fixed dimensional vector summarising the representations in the shared layers. The summary vector is passed through a fully connected layer followed by a $\softmax$  to predict the probability distribution over the tasks:
\begin{eqnarray*}
	P_{\vTheta_d}(\textrm{task id}|\vh_d) \sim \softmax (\vW_d \vh_d + \vb_d) \\
	\vh_d := \dlstm(\srep_{\vTheta_{mtl}}(\vx,\vy))
\end{eqnarray*}
where $\dlstm$ denotes the discriminator LSTMs, $\srep_{\vTheta_{mtl}}(\vx,\vy)$ denotes the representations in the shared layer of deep encoders and decoders in the MTL architecture, and $\vTheta_d$ includes the $\dlstm$ parameters as well as $\{\vW_d,\vb_d\}$.

\paragraph{Adversarial Objective.}
Inspired by \cite{chen2017adversarial}, we add two additional terms to the MTL training objective in eqn \ref{mtl_obj}. The first term is $\mathcal{L}_{adv1}(\vTheta_d) $ defined as:

\vspace{-5mm}
{\small
\begin{eqnarray*}
	\sum_{m=0}^M \sum_{(\vx,\vy) \in \mathcal{D}_m} \log P_{\vTheta_d}(m | \dlstm(\srep_{\vTheta_{mtl}}(\vx,\vy))).
\end{eqnarray*} 
}
Maximising the above objective over $\vTheta_d$ ensures proper training of the discriminator to predict the identity of the task. 
The second term ensures that the parameters of the shared layers are trained so that they confuse the discriminator by maximising the entropy of its predicted distribution over the task identities. That is, we add the term $\mathcal{L}_{adv2}(\vTheta_{mtl})$ to the training objective  defined as: 

{\small
\begin{equation*}
	\sum_{m=0}^M \sum_{(\vx,\vy) \in \mathcal{D}_m} H\big[P_{\vTheta_d}(. | \dlstm(\srep_{\vTheta_{mtl}}(\vx,\vy)) )\big]
\end{equation*}
}
where $H[.]$ is the entropy of a distribution. In summary, the adversarial training leads to the following optimisation
\begin{equation*}
\argmax_{\vTheta_d,\vTheta_{mtl}}  \mathcal{L}_{mtl}(\vTheta_{mtl}) + \mathcal{L}_{adv1}(\vTheta_d)  + \lambda \mathcal{L}_{adv2}(\vTheta_{mtl}).
\end{equation*}
We maximise the above objective by SGD, and update the parameters by alternating between optimising $\mathcal{L}_{mtl}(\vTheta_{mtl})+\lambda \mathcal{L}_{adv2}(\vTheta_{mtl})$ and $\mathcal{L}_{adv1}(\vTheta_d)$.

%% file: sec60-experiments.tex
\subsection{Bilingual Corpora}
We use three language-pairs, translating from English to French, Farsi, and Vietnamese. We have chosen these languages to \reza{analyse} the effect of multi-task learning on languages with different underlying linguistic structures. The sentences are segmented using BPE \cite{sennrich-haddow-birch:2016:P16-12} on the union of source and target vocabularies for English-French and English-Vietnamese. For English-Farsi, BPE is performed using separate vocabularies due to the disjoint alphabets.
We use a special  \unk token to replace  unknown BPE units in the test and development sets. 
 
Table \ref{tab:corp-stat} show some statistics about the bilingual corpora. \reza{Further details about the corpora and their pre-processing is as follows:} 
\begin{itemize}[leftmargin=5mm]
\item  The English-French corpus \poorya{is a random subset of EuroParlv7 as distributed to WMT2014}.  Sentence pairs in which either the source or the target has length more than 80 (before applying BPE) have been removed.  The BPE is performed with a 30k total vocabulary size.  The ``news-test2012'' and ``news-test-2013'' portions are used for validation and test sets, respectively.

\item The English-Farsi corpus is assembled from all the parallel news text in LDC2016E93 \emph{Farsi Representative Language Pack} from the Linguistic Data Consortium, combined with English-Farsi parallel subtitles from the TED corpus \cite{tiedemann2012opus}.  
Since the TED subtitles are user-contributed, this text contained considerable variation in the encoding of its Perso-Arabic characters.  To address this issue, we have normalized the corpus using the Hazm toolkit\footnote{\texttt{www.sobhe.ir/hazm}}.
Sentence pairs in which one of the sentences has more than 80  (before applying BPE) are removed, and BPE is performed with a 30k vocabulary size.  
Random subsets of this corpus (3k and 4k sentences each) are held out as validation and test sets, respectively.

\item The English-Vietnamese is from the translation task in IWSLT 2015, and we use the preprocessed version provided by \cite{Luong-Manning:iwslt15}. The sentence pairs in which at least one of their sentences had more than 300 units (after applying BPE) are removed. ``tst2012" and ``tst2013" parts are used for validation and test sets, respectively.
\end{itemize}

\input{table2.tex}

\subsection{Auxiliary Tasks}
We have chosen the following auxiliary tasks to provide the NMT model with syntactic and/or semantic knowledge, in order to   enhance the quality of translation:

\paragraph{Named-Entity Recognition (NER).} With a small bilingual training corpus, it would be hard for the NMT model to learn how to translate rarely occurring named-entities. Through the NER  task, the model hopefully learns the skill to recognize named entities. 
Speculatively, it would then enables leaning translation patterns by masking out named entities. 
The NER data comes from the CONLL shared task.\footnote{https://www.clips.uantwerpen.be/conll2003/ner}

\paragraph{Syntactic Parsing.} This task enables NMT to learn the phrase structure of the input sentence, which would then be useful in better re-orderings. This would be most useful for language pairs with high syntactic divergence. The parsing data comes from the Penn Tree Bank with the standard split for training, development, and test \cite{Marcus:1993}. 
We linearise the constituency trees, in order to turn syntactic parsing as a \sq transduction \cite{NIPS2015_5635}.  

\paragraph{Semantic Parsing.} A good translation should preserve the meaning. 
Learning from the  semantic parsing task enables the NMT model to pay attention to a meaning abstraction of the source sentence, in order to  convey it to the target translation. 
We have made use of the Abstract Meaning Representation (AMR) corpus Release 2.0 (LDC2017T10), which pairs English sentences AMR meaning graphs. 
We linearise the AMR graphs, in order to convert semantic parsing as a \sq transduction problem \cite{amr_parsing17}.

\input{table1.tex}
\subsection{Models and Baselines}

We have implemented the proposed multi-task learning architecture in C++ using DyNet  \cite{2017arXiv170103980N}, on top of 
Mantis \cite{cohn-EtAl:2016:N16-1} which is an implementation of the attentional \sq NMT model in \cite{bahdanau:ICLR:2015}.
In our multi-task architecture, we do partial sharing of parameters, where  the parameters of the top 2 stacked layers are shared among the encoders of the tasks. Moreover, we share the parameters of the top layer stacked decoder among the tasks. 
\reza{Source and target embedding tables are shared among the tasks, while the attention component is task-specific. \footnote{In our experiments, models with task-specific attention components achieved better results than those sharing them.}}
We compare against the following baselines:
\begin{itemize}[leftmargin=5mm]
\item Baseline 1: The vanila \sq model without any multi-tasking. 
\item Baseline 2: The multi-tasking architecture proposed in \cite{niehues2017exploiting}, which is a special case of our approach where the parameters of all 3 stacked layers are shared among the tasks.\footnote{We have used their best performing architecture and changed the training schedule to ours.} They have not used deep stacked layers in encoder and decoder as we do, so we extend their work to make it  comparable with ours.
\end{itemize}
The configuration of models is as follows.   The encoders and decoders make use of GRU units with 400 hidden dimensions, and the attention component has  200 dimensions. 
 For training, we used Adam algorithm \cite{kingma2014adam} with the initial learning rate of 0.003 for all of the tasks. Learning rates are halved when the performance on the corresponding dev set decreased. In order to speed-up the training, we use mini-batching with the  size of 32. Dropout rates for both encoder and decoder are set to 0.5, and models are trained for 50 epochs where the best models is selected based on the perplexity on the dev set. $\lambda$ for the adversarial training is set to 0.5. 
Once  trained, the NMT  model translates using the greedy search. We use BLEU \cite{Papineni:2002:BMA:1073083.1073135} to measure translation quality. \footnote{With ``multi-bleu.perl'' script from Moses \cite{Koehn:2007:MOS:1557769.1557821}.}

\subsection{Results} 
Table \ref{tab:main-results-full} reports the  BLEU scores and perplexities for the baseline and our proposed method on the three aforementioned translation tasks.  
It can be seen that the performance of multi-task learning models are better than Baseline 1 (only MT task). This confirms that adding auxiliary tasks  helps to increase the performance of the machine translation task. 

As expected, the effect of different tasks are not similar across the language pairs, possibly due to the following reasons: 
(i) these translation tasks datasets come from different domains so they have various degree of domain relatedness to the auxiliary tasks, 
and (ii) the BLEU scores of the Baseline 1 show that the three translation models are on different quality levels  which may entail that they benefit 
from auxiliary knowledge on different levels. 
In order to improve a model with low quality translations due to language divergence, syntactic knowledge can be more helpful as they help better reorderings. 
In a higher-quality model, however, semantic knowledge can be more useful as a higher-level linguistic knowledge. 
This pattern can be seen in the reported results: syntactic parsing leads to more improvement on Farsi  translation which has a low BLEU score and high language divergence to English, and semantic parsing yields more improvement on the Vietnamese translation task which already has a high BLEU score. 
The NER task has led to a steady improvement in all the translation tasks, as it leads to better handling of named entities. 
 
We have further added adversarial training to ensure the shared representation learned by the encoder is not contaminated by the task-specific information. 
The results are in the last row of Table  \ref{tab:main-results-full}. The experiments show that adversarial training leads to further gains in  MTL translation quality, except when translating into Farsi. We speculate this is due to the low quality of NMT for Farsi, 
where  updating shared parameters with respect to the entropy of discriminator's predicted distribution may negatively affect the model.

\input{table3.tex}
Table \ref{tab:baseline2} compares our multi-task learning approach to Baseline 2. 
As Table \ref{tab:baseline2}, our partial parameter sharing mechanism is more effective than fully sharing the parameters (Baseline 2), due to its flexibility in allowing access to  private task-specific knowledge.
We also applied the adaptation technique \cite{niehues2017exploiting} as follows. Upon finishing MTL training, we  continue to train only on the MT task for another 20 epochs, and choose the best model based on perplexity on dev set. 
Adaptation has led to consistent gains in the performance of our MTL architecture and Baseline 2.

%% file: table2.tex
\begin{table}[]
\centering
\begin{tabular}{c|c|c|c}
& Train  & Dev  & Test \\
\hline
En $\rightarrow$ Fr & 98,846  & 5,357 & 5,357 \\
\hline
En $\rightarrow$ Fa & 98,158  & 3,000 & 4,000 \\
\hline
En $\rightarrow$ vi & 133,290 & 1,553 & 1,268 
\end{tabular}
\caption{The statistics of bilingual corpora.}
\label{tab:corp-stat}
\end{table}

%% file: table1.tex
\setlength\tabcolsep{4pt}
\begin{table*}[t]
\small
\begin{minipage}{\textwidth}
\centering
\begin{tabular}{|l||c|c|c|c||c|c|c|c||c|c|c|c|}

\multicolumn{1}{c}{ } & \multicolumn{4}{c}{English $\rightarrow$ French} & \multicolumn{4}{c}{English $\rightarrow$ Farsi} & \multicolumn{4}{c}{English $\rightarrow$ Vietnamese} \\
\cline{2-13}
\multicolumn{1}{c|}{ } & \multicolumn{2}{c|}{Dev} & \multicolumn{2}{c|}{Test} & \multicolumn{2}{c|}{Dev} & \multicolumn{2}{c|}{Test} & \multicolumn{2}{c|}{Dev} & \multicolumn{2}{c|}{Test}  \\
\multicolumn{1}{c|}{}  &  PPL & BLEU & PPL & BLEU & PPL & BLEU &  PPL & BLEU & PPL & BLEU & PPL & BLEU  \\
\hline \hline
 NMT  & 117.27 & 8.85 & 80.29 & 10.71 & 86.63 & 7.69 & 87.94 & 7.46 & 23.24 & 16.53 & 20.36 & 17.86 \\
\ \ \ + Semantic & 71.7 & 10.58 & 51.2 & 12.72 & 56.32 & 8.3 & 57.88 & 8.32 & 14.86 & 19.96 & 12.79 & 21.82 \\
\ \ \ + NER & 73.42 & 10.73 & 52.07 & 12.92 & 48.46 & 9.11 & 49.53 & 9.03 & 15.04 & 20.2 & 13.13 & 21.96 \\
\ \ \  + Syntactic & 69.45 & 11.88 & 48.9 & 13.94 & 44.35 & 9.73 & 45.37 & 9.37 & 16.42 & 18.4 & 14.27 & 20.4 \\
\ \ \ + All Tasks & 69.71 & 11.3 & 49.86 & 13.41 & 44.03 & \textbf{9.68} & 45.1 & \textbf{9.7} & 14.79 & 20.12 & 12.65 & 22.41 \\
\ \ \ + All+Adv. & 68.44 & \textbf{11.93} & 48.92 & \textbf{14.02} & 45.25 & 9.55 & 45.87 & 9.19 & 14.19 & \textbf{21.21} & 12.11 & \textbf{23.54} \\
\hline
\end{tabular}
\caption{BLEU scores and perplexities of the baseline vs our MTL architecture with various auxiliary tasks on the full bilingual datasets.}
\label{tab:main-results-full}
\end{minipage}
\end{table*}

%% file: table3.tex
\setlength\tabcolsep{3.5pt}
\begin{table}[]
\centering
{\small
\begin{tabular}{l||c|c||c|c|c|}
 & \multicolumn{2}{c||}{\poorya{W/O Adaptation}} & \multicolumn{3}{c}{\poorya{W/ Adaptation}} \\
& Partial  & Full  & Partial & \poorya{Part.+Adv.} & Full\\
\hline \hline
En$\rightarrow$Fr  &  13.41 & 9.94 & 14.86 & \poorya{15.12} & 11.94 \\
En $\rightarrow$ Fa & 9.7 & 7.89 & 10.31 & \poorya{10.08} & 8.6\\
En $\rightarrow$ Vi  & 22.41 & 20.26 & 23.35 & \poorya{24.28} & 21.67\\
\end{tabular}
}
\caption{Our method (partial parameter sharing) against Baseline 2 (full parameter sharing).}
\label{tab:baseline2}
\end{table}

%% file: sec70-analysis.tex
\begin{figure*}[t]
\centering
\includegraphics[scale=0.5]{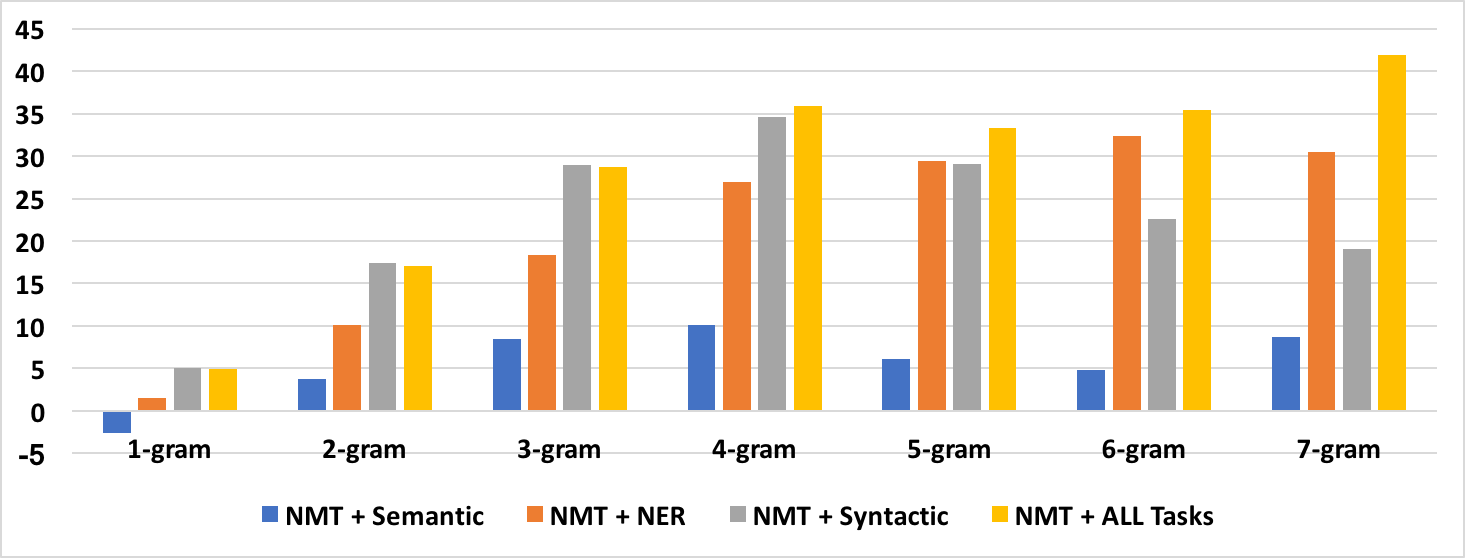}
\caption{Percentage of more correct $n$-grams generated by the deep MTL models compared to the single-task model (only MT).  }
\label{fig:ngrams}
\end{figure*} 

\paragraph{How many layers of encoder/decoder to share?}  
Figure \ref{fig:share-enc-layers}  show the results of changing the number of shared layers in encoder and decoder based on the En$\rightarrow$Vi translation task. The results confirm that partial sharing of stacked layers is better than full sharing. Intuitively, partial sharing provides the model with an opportunity to learn task specific skills via the private layers, while leveraging the knowledge learned from other tasks via shared layers. 

\input{fig_layers_analysis.tex}

\paragraph{Statistics of gold $n$-grams in MTL translations.} Generating high order gold  $n$-grams is hard. We analyse the effect of syntactic and semantic knowledge on generating gold $n$-grams in translations. 

For each sentence, we first extract $n$-grams in the gold translation, and then compute the number of $n$-grams which are common with the generated translations.  Finally, after aggregating the results over the entire test set,  we compute the percentage of additional gold $n$-grams generated by each MTL model compared to  the ones in single-task MT model. The results are depicted in Figure \ref{fig:ngrams}. Interestingly, the MTL models generate more correct $n$-grams relative to the vanilla NMT model,  as $n$ increases.

\input{table_examples.tex}

\paragraph{Effect of the NER task.} 
The NMT model has difficulty translating rarely occurring named-entities, particularly when the bilingual parallel data is scarce. 
We expect  learning from the NER task leads the MTL model to recognize named-entities and learn underlying patterns for translating them. 
The top part in Table \ref{tab:examples} shows an example of such situation.  As seen, the MTL is able to recognize all of the named-entities in the sentence and translate the while the single-task model missed ``India''. 

For more analysis, we have applied a Farsi POS tagger \cite{feely2014cmu} to gold translations. Then, we extracted $n$-grams with at least one noun in them, and report the statistics of correct such $n$-grams, similar to what reported in  Figure \ref{fig:ngrams}. 
The resulting statistics  is depicted in Figure \ref{fig:noun_grams}.
As seen, the MTL model trained on  MT and NER tasks leads to generation of  more correct unigram noun phrases relative to the vanilla NMT,  as $n$ increases. 

\input{fig_noun_ngrams.tex}

\paragraph{Effect of the semantic parsing task.}
Semantic parsing encourages a  precise understanding of the source text, which would then be useful for conveying the correct meaning to the translation. 
The middle part in Table \ref{tab:examples} is an example translation,  showing that semantic parsing has helped NMT by understanding  that ``the subject sees the object via subject's screens''.

\paragraph{Effect of the syntactic parsing task.} Recognizing the syntactic structure of the source sentence helps NMT to better translate  phrases. The bottom part of  Table \ref{tab:examples} shows an example translation demonstrating such case.  
The source sentence is talking about ``a method for controlling the traffic'', which is correctly translated by the MTL model while  vanilla NMT has mistakenly translated it to ``controlled traffic''.

%% file: fig_layers_analysis.tex
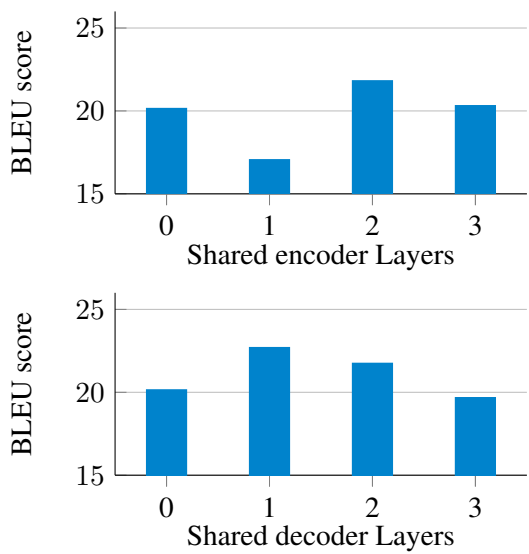
\begin{figure}[!h]
	\centering
\begin{tabular}{c}
	\begin{tikzpicture} 
		\begin{axis}[
			ybar,
			xlabel = Shared encoder Layers,
			xmin = -0.5,
			xmax = 3.5,
			ymin = 15,
			ymax = 26,
			axis x line* = bottom,
			axis y line* = left,
			ylabel= BLEU score,
			width= 7cm,
			height= 4cm,
			ymajorgrids = true,
			bar width = 5mm,
			xticklabels = \empty,
			extra x ticks = {0,1,2,3,4},
			extra x tick labels = {0,1,2,3,4},
			]
			\addplot+[mark=none, blau_2b, very thick] coordinates {
				(0,20.09)
				(1,17)
				(2,21.76)
				(3,20.26)
			};
		\end{axis} 
	\end{tikzpicture}
\\
	\begin{tikzpicture} 
			\begin{axis}[
			ybar,
			xlabel = Shared decoder Layers,
			xmin = -0.5,
			xmax = 3.5,
			ymin = 15,
			ymax = 26,
			axis x line* = bottom,
			axis y line* = left,
			ylabel= BLEU score,
			width= 7cm,
			height= 4cm,
			ymajorgrids = true,
			bar width = 5mm,
			xticklabels = \empty,
			extra x ticks = {0,1,2,3,4},
			extra x tick labels = {0,1,2,3,4},
			]
			\addplot+[mark=none, blau_2b, very thick] coordinates {
				(0,20.09)
				(1,22.64)
				(2,21.69)
				(3,19.62)
			};
		\end{axis} 
	\end{tikzpicture}
\end{tabular}
		
	\caption{BLEU scores for different numbers of shared layers in (top)  encoder while no layer is shared in decoder,
	and (bottom)  decoder while no layer is shared in encoder }
	\label{fig:share-enc-layers}
\end{figure}

%% file: table_examples.tex
\begin{table*}[!t]
\centering
\renewcommand{\tabularxcolumn}{m} 
{\small
\begin{tabularx}{\textwidth}{|c|X|}
\hline
 English & this is a building in Hyderabad , \underline{India} .   \\
  Reference & this a building in Hyderabad is , in \underline{India} .\\
 MT only model & this a building in Hyderabad is .\\
 MT+NER model & this a building in Hyderabad \underline{India} is .\\
\hline
 English & we see people on our screens .\tabularnewline
  Reference & we people \underline{on television screen} or cinema see .  \tabularnewline
  MT only model & we people see we people .  \tabularnewline
  MT+semantic model & we people \underline{on television screen} see . \tabularnewline
\hline

 English & in hospitals , for new medical instruments ; in streets for traffic control . \tabularnewline
  Reference & in hospitals , \underline{for instruments medical new} ; in streets for \underline{control traffic}  \tabularnewline
  MT only model & in hospitals , \underline{for tools new tools for traffic controlled}* \footnote{it mistakenly translated "control traffic" to "controlled traffic"} .\tabularnewline
  MT+syntactic model & in hospitals , \underline{for devices new} , in streets for \underline{control traffic} . \tabularnewline
\hline
\end{tabularx}
}
\caption{Example of translations on Farsi test set. In this examples each Farsi word is replaced with its English translation, and the order of words is reversed (Farsi is written right-to-left). The structure of Farsi  is Subject-Object-Verb (SOV), leading to  different word orders in English and Reference sentences.}
\label{tab:examples}
\end{table*}

%% file: fig_noun_ngrams.tex
\begin{figure}
	\raggedright 
	\small
	\begin{tikzpicture} 
	\tikzstyle{every node}=[font=\small]
			\begin{axis}[
			ybar,			xmin = -0.5,
			xmax = 6.5,
			ymin = 0,
			ymax = 40,
			axis x line* = bottom,
			axis y line* = left,
			width= 8cm,
			height= 5cm,
			ymajorgrids = true,
			bar width = 5mm,
			xticklabels = \empty,
			extra x ticks = {0,1,2,3,4,5,6},
			xticklabel style={rotate=90},
			extra x tick labels = {1-gram,2-gram,3-gram,4-gram,5-gram,6-gram,7-gram,8-gram,9-gram, 10-gram},
			]
			\addplot+[mark=none, blau_2b, very thick] coordinates {
				(0,1.78966207)
				(1,9.321192053)
				(2,17.02422145)
				(3,25.4857997)
				(4,27.89317507)
				(5,31.32183908)
				(6,31.11111111)
			};
		\end{axis} 
	\end{tikzpicture}
	\caption{Percentage of more corrected n-grams with at least one noun generated by MT+NER model compared with the only MT model (only MT).  }\label{fig:noun_grams}
\end{figure}
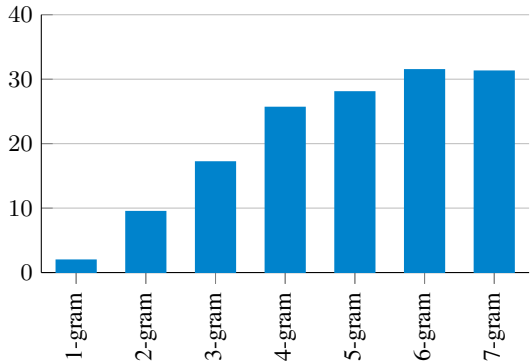

%% file: sec80-related.tex
Multi-task learning has attracted attention to  improve NMT in recent work. 
\cite{zhang2016exploiting} has made use of monolingual data in the source language in a  multitask learning framework  by sharing encoder in the attentional encoder-decoder model. 
Their auxiliary task is to reorder the source text to make it close to the target language word order. 
\cite{domhan2017using} proposed a two-layer stacked decoder, which the bottom layer is trained on language modelling on the target language text.  
The next word is jointly predicted by the bottom layer language model and the top layer attentional RNN decoder.
They reported only moderate improvements over the baseline and fall short against using synthetic parallel data.
\poorya{
\cite{dalvi2017understanding} investigated the amount of learned morphology and how it can be injected using MTL.
Our method is related to what they call “joint data-learning”, where they share all of the \sq components among the tasks. 
}

 \poorya{
\cite{belinkov2017neural, shi2016does, belinkov2017evaluating} investigate syntax/semantics phenomena learned as a byproduct of \sq NMT training. 
We, in turn, investigate the effect of “injecting” syntax/semantic on learning NMT using MTL.
}

The closet work to ours is \cite{niehues2017exploiting}, which has made use of  part-of-speech tagging and named-entity recognition tasks to improve NMT. They have used the attentional encoder-decoder with a shallow architecture, and share different parts eg the encoder, decoder, and attention. They report the best performance with fully sharing  the encoder. In contrast, our architecture uses partial sharing on deep stacked encoder and decoder components, and the results show that it is critical for  NMT improvement in MTL. Furthermore, we propose adversarial training to prevent contamination of shared knowledge with task specific details. 

Taking another approach to MTL,  \cite{sogaard2016deep} and \cite{hashimoto2016joint} have proposed  architectures by stacking up tasks 
on top of each other according to their linguistic level, eg from lower level tasks (POS tagging) to higher level tasks (parsing). 
In this approach, each task uses predicted annotations and hidden states of the lower-level tasks for making a better prediction. 
This is contrast to the approach taken in this paper where models with shared parameters are trained jointly on multiple tasks. 

More broadly,  deep multitask learning has been used for various NLP problems, including graph-based parsing \cite{2011arXiv1101.6081C} and 
keyphrase boundary classification  \cite{DBLP:conf/acl/AugensteinS17} .  \cite{chen2017adversarial} has applied multi-task learning for Chinese word segmentation, and \cite{liu2017adversarial} applied it for text classification problem. Both of these works have used adversarial training to make sure the shared layer extract only common knowledge. 

MTL has been used effectively to learn from multimodal data. \cite{iclr44928} has proposed MTL  architectures for neural \sq transduction for  tasks including MT, image caption generation, and parsing. 
They  fully share the encoders (many-to-one), the decoders (one-to-many), or some of the encoders and decoders (many-to-many). 
\cite{pasunuru2017multitask} \reza{have} made use of an MTL approach to improve video captioning with auxiliary tasks including video prediction and logical language entailment based on a many-to-many architecture.

%% file: sec90-conc.tex
We have presented an approach to improve NMT in bilingually scarce scenarios, by 
leveraging curated linguistic resources in the source, including semantic parsing, syntactic parsing, and named entity recognition.
This is achieved via an effective MTL architecture, based on deep stacked encoders and decoders, to share common knowledge among the MT and auxiliary tasks. 
Our experimental results show substantial improvements in the translation quality, when translating from English to French, Vietnamese, and Farsi in bilingually scarce scenarios. 
For future work, we would like to investigate architectures which allow automatic parameter tying among the tasks \cite{DBLP:journals/corr/RuderBAS17}.